%% file: root.tex
\documentclass[conference]{IEEEtran}
\IEEEoverridecommandlockouts
\usepackage{cite}
\usepackage{amsmath,amssymb,amsfonts}
\usepackage{algorithmic}
\usepackage{graphicx}
\usepackage{textcomp}
\usepackage{xcolor}
\usepackage{siunitx}
\usepackage{pgfplots}
\pgfplotsset{compat=1.18}
\usetikzlibrary{plotmarks}
\usepgfplotslibrary{groupplots}
\usepgfplotslibrary{statistics}
\usepgfplotslibrary{external}
\usepackage[inkscapearea=page]{svg}
\usepackage{adjustbox}
\usepackage{placeins}
\usepackage{hyperref}
\usepackage[capitalize]{cleveref}
\usepackage{subfigure}
\usepackage{booktabs}
\usepackage{makecell}
\usepackage{tabularx}

\def\BibTeX{{\rm B\kern-.05em{\sc i\kern-.025em b}\kern-.08em
    T\kern-.1667em\lower.7ex\hbox{E}\kern-.125emX}}

\begin{document}

\definecolor{TUMGreen}{RGB}{73, 156, 0} % {148, 255, 105}
\definecolor{TUMBlue}{RGB}{0,101,189}
\definecolor{TUMOrange}{RGB}{227, 114, 34}

\title{\LARGE \bf Investigating Driving Interactions: A Robust Multi-Agent Simulation Framework for Autonomous Vehicles}

\author{Marc Kaufeld$^{1,*}$, Rainer Trauth$^{2,*}$ and Johannes Betz$^{1}$
\thanks{$^{1}$ M. Kaufeld, J. Betz are with the Professorship of Autonomous Vehicle Systems, Technical University of Munich, 85748 Garching, Germany; Munich Institute of Robotics and Machine Intelligence (MIRMI).}
\thanks{$^{2}$ The author is with the Institute of Automotive Technology, Technical University of Munich, 85748 Garching, Germany; Munich Institute of Robotics and Machine Intelligence (MIRMI). They gratefully acknowledge the financial support from the Technical University of Munich and the Bavarian Research Foundation (BFS)}
\thanks{$^{*}$Shared first authorship.}
}

\maketitle

\begin{abstract}
Current validation methods often rely on recorded data and basic functional checks, which may not be sufficient to encompass the scenarios an autonomous vehicle might encounter.
In addition, there is a growing need for complex scenarios with changing vehicle interactions for comprehensive validation.
This work introduces a novel synchronous multi-agent simulation framework for autonomous vehicles in interactive scenarios.
Our approach creates an interactive scenario and incorporates publicly available edge-case scenarios wherein simulated vehicles are replaced by agents navigating to predefined destinations. 
We provide a platform that enables the integration of different autonomous driving planning methodologies and includes a set of evaluation metrics to assess autonomous driving behavior.
Our study explores different planning setups and adjusts simulation complexity to test the framework's adaptability and performance. 
Results highlight the critical role of simulating vehicle interactions to enhance autonomous driving systems.  
Our setup offers unique insights for developing advanced algorithms for complex driving tasks to accelerate future investigations and developments in this field.
The multi-agent simulation framework is available as open-source software: \\\url{https://github.com/TUM-AVS/Frenetix-Motion-Planner}
\end{abstract}


\section{Introduction}
\begin{figure}[!t]%
	\centering%
	  \includegraphics[width=0.9\linewidth]{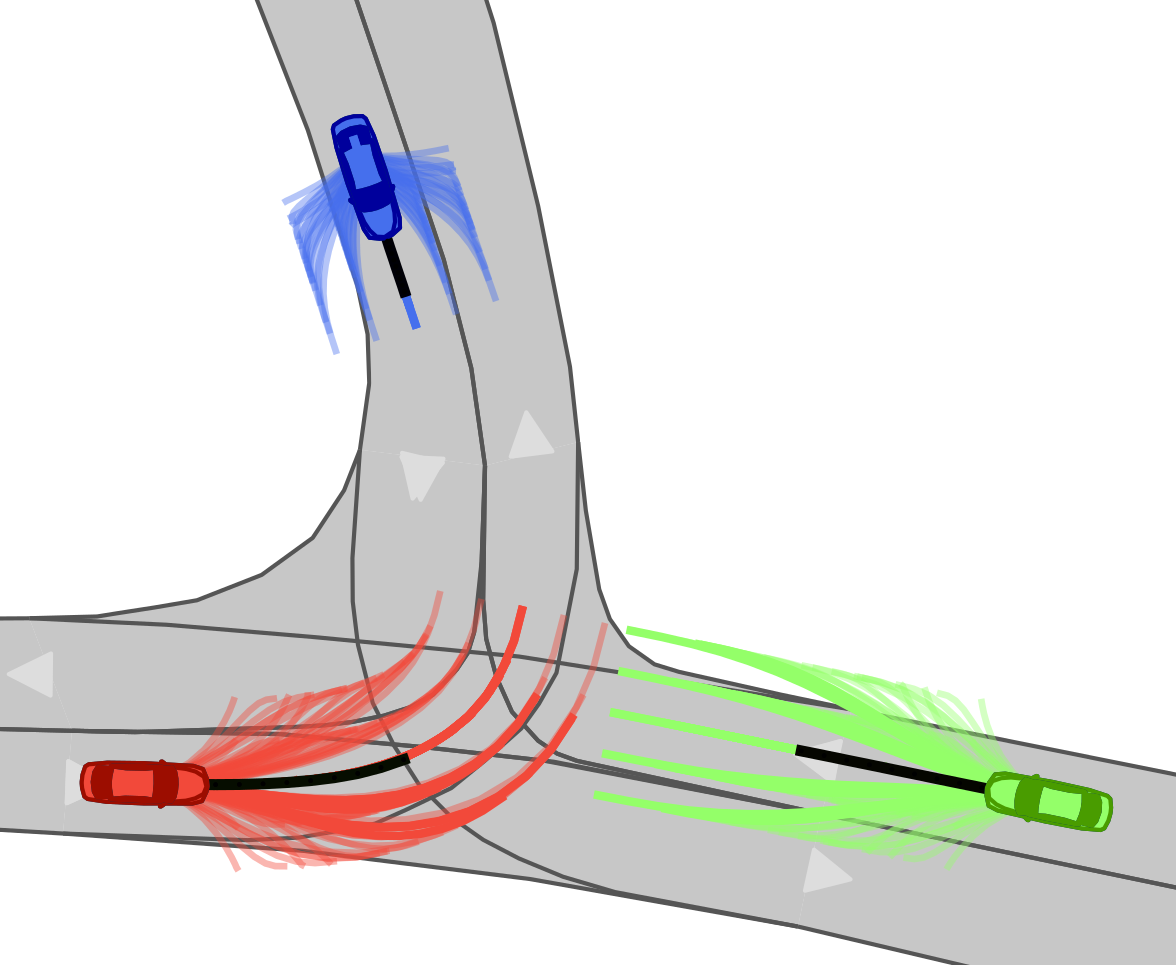}
	\caption{Illustrative example of the multi-agent simulation framework with three agents, each controlled by a sampling-based motion planner.}%
	\label{fig:intro}
\end{figure}

Testing over several billion kilometers is required to statistically demonstrate that autonomous vehicles (AVs) are as safe as human drivers regarding fatal accidents~\cite{Wachenfeld2015}. However, such real-world demonstrations are economically infeasible~\cite{Riedmaier.2020}. For this reason, the simulation of AVs in various scenarios is essential for accelerating the development and evaluation of software for autonomous driving. 
Nevertheless, most simulation environments consider interactions between several road users only to a limited extent.
While the agent vehicle can adapt to surrounding changes, the behavior of other vehicles does not necessarily provide a similar responsiveness.
However, the reciprocal influence of interactions substantially shapes the collective behavior of traffic participants.
Consequently, it is necessary to recognize the interplay among multiple road users~(\cref{fig:intro}). 
To address these shortcomings, this paper presents a multi-agent simulation framework to address complex scenarios for AVs in reactive environments. The proposed simulation framework is set up to facilitate numerous agents to reach their designated goal regions concurrently.
In summary, this work provides the following contributions:
\begin{itemize}
    \item We present an advanced multi-agent simulation framework compatible with the CommonRoad benchmark scenarios~\cite{Althoff.2017} for the reproducible simulation of interactive driving scenarios for AVs.
    \item An interface is provided to integrate and compare different trajectory planning algorithms for AVs, including metrics to evaluate vehicle behavior.
    \item We evaluate the framework's ability to analyze agent behavior and its single- and multi-core computational performance.
\end{itemize}

\section{Related Work}
There are several simulation environments for AVs in the current state of the art. An overview is given by Tong et al.~\cite{Tong.2020}. 
Three-dimensional simulation environments such as Carla~\cite{Dosovitskiy.2017}, CarSim ~\cite{carsim} or Apollo~\cite{apollo} focus on detailed physics and high-quality rendering for realistic motion description and detailed visualization of individual vehicles. Similarly, Gazebo~\cite{Koenig.2004} is a built-in simulation software for ROS that can evaluate AV driving software in three-dimensional environments. These frameworks are frequently used to assess the complete AV software stack from sensor input to actuator output. They often allow high-quality modeling of sensors, making them well-suited for evaluating and testing the holistic AV functionality. 
Yet, generating complex and interactive scenarios to evaluate different trajectory planning methods is effortful.
Simulation environments must constitute appropriate reproducible scenarios to test and compare different trajectory planning algorithms. 
Numerous benchmarking frameworks for navigation tasks in grid- or polygon-based environments exist in robotics, e.g., Moving AI~\cite{Sturtevant.2012} or Bench-MR~\cite{Heiden.2021}.
However, a detailed street network is necessary to simulate trajectory planning for autonomous driving. 
CommonRoad~\cite{Althoff.2017} was amongst the first environments providing reproducible benchmark scenarios. 
It offers recorded and hand-crafted driving scenarios in a bird's-eye view.
Since the hand-crafted scenarios are optimized for a minimal driveable area, CommonRoad tackles the necessity of providing critical edge-case scenarios. Still, they are not interactive, and traffic does not react to the agent's behavior. 
Scenarios with large-scale traffic flow simulations can be created using traffic simulators such as SUMO~\cite{Lopez.2018} or SimMobilityST~\cite{Azevedo.2017}. Both employ simple rule-based behavior models such as the IDM (Intelligent Driver Model)~\cite{Treiber.2017} to generate fundamental vehicle interactions but do not aim at an accurate motion description of individual vehicles.
Since the interaction between traffic participants is crucial for reproducing the complexity of real-world traffic conditions, an extension to CommonRoad is proposed in~\cite{Klischat.2019} including SUMO, where the traffic simulator controls the simulated vehicular movement. Although artificial vehicles can react to the agent's motion by accelerating or decelerating, complex actions and maneuvers of the simulated vehicles cannot be modeled.
Similarly, nuPlan~\cite{Caesar.2021} provides recorded scenarios with a non-interactive environment. They also include simple planning models that simulate anticipating traffic flow. 
None of these benchmark environments encompass entirely reactive traffic participants with complete freedom for trajectory planning.
Only a few multi-agent simulation frameworks feature interactive environments where the controlled vehicle can respond to surrounding vehicles and vice versa.
TORCS (The open racing car simulator)~\cite{Torcs.2000} can be used to test planning algorithms for autonomous racing with multiple vehicles.
CoInCar-Sim~\cite{Naumann.2018} serves as a framework for simulating cooperative AVs. The authors present this framework as a benchmark for cooperative planning algorithms. The simultaneous behavioral planning for numerous agents is considered while assuming information exchange through V2X communication. The emphasis is put on simulating and comparing collaborative planning algorithms. 
Bernhard et al.~\cite{Bernhard.2020} propose BARK (Behavior BenchmARK) as a multi-agent simulation that replaces recorded vehicles in open-source data sets like INTERACTION~\cite{Zhan.2019} with agents controlled by distinct behavioral planning models. They incorporate analytical methods such as the IDM~\cite{Treiber.2017} or MOBIL (Minimizing Overall Braking Induced by Lane change) models~\cite{Kesting.2007}, but also reinforcement learning approaches. Through simultaneous planning, every agent can react to nearby vehicles. 
Yet, simulations based on recorded data often lack edge-case scenarios to assess critical situations. Besides, only basic measures are provided to evaluate the planning methods. The included metrics are limited to individual agents' success or failure, but the overall criticality of the scenarios due to the agents' decisions is not assessed.
Finally, regarding real-world interactive AV behavior, Kloock et al.~\cite{Kloock.2021} use remote-controlled model cars on a test track for a small-scale, real-world multi-vehicle simulation with up to twenty vehicles. 
The vehicles can be controlled either centralized, simulating cooperative behavior, or through distributional computation to conduct experiments with non-cooperative agents.

\section{Methodology of the Multi-agent Simulation}
\begin{figure*}[!htbp]%
	\centering%
	\includegraphics[width=\linewidth, trim=1.5cm 13.5cm 1.5cm 17.7cm, clip]{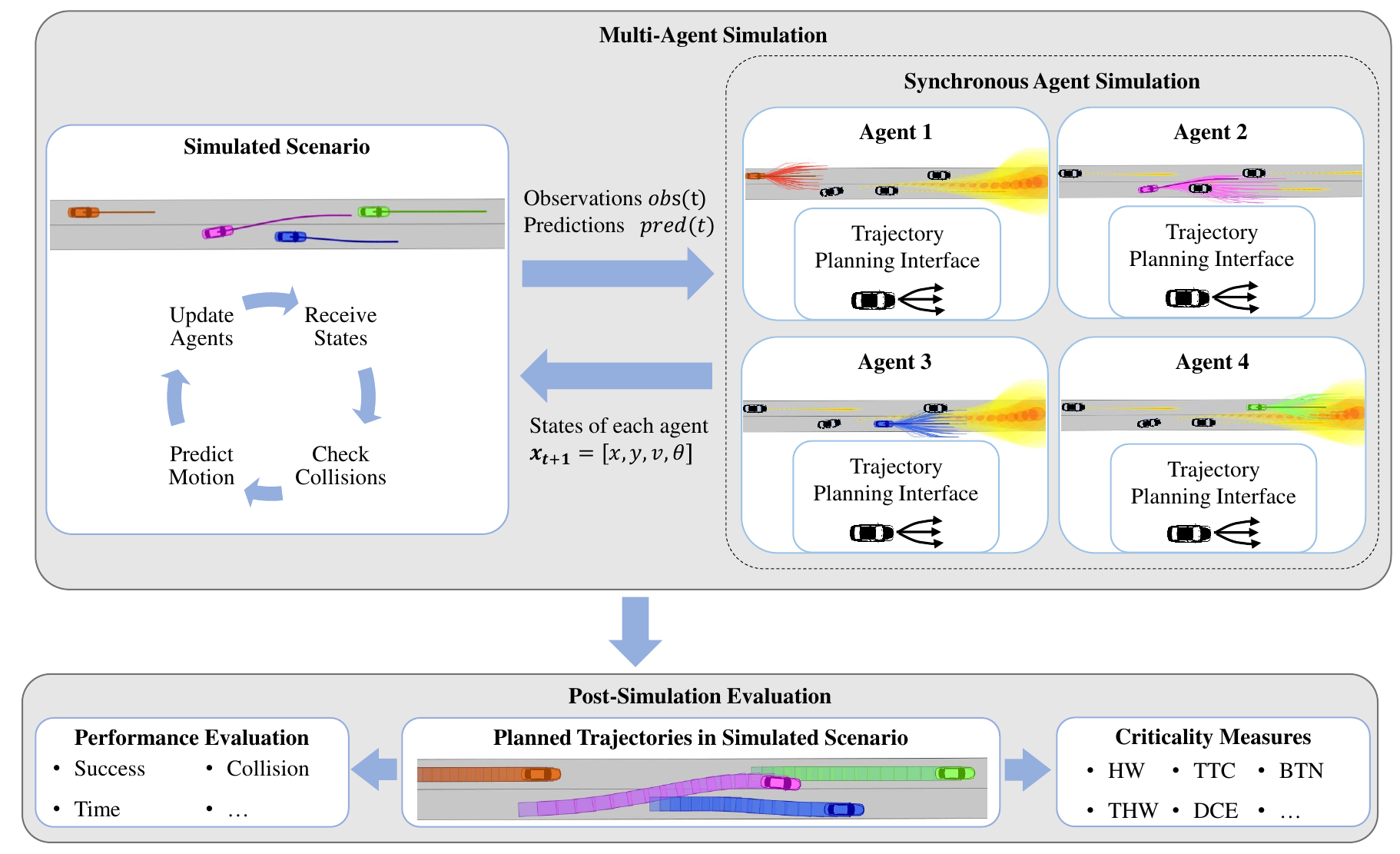}
	\caption{The multi-agent simulation framework structure illustrated with a scenario involving four agents: In each time step, the states of all agents are aggregated in the global scenario, where a collision check is conducted. Subsequently, the future movement of all vehicles is predicted, and the information is shared with each agent. Every agent maintains a local representation of the scenario, including only the positions and predictions of surrounding vehicles. Based on this information, agents calculate their next trajectory step.
  The yellow and orange areas in the figures of each agent show the probability-based motion predictions, while the spawned trajectories are the sampled paths to select the next motion step. In the global scenario, every agent follows their individual best trajectory.
    When the simulation is finished, the final trajectories are evaluated with criticality measures.}%
	\label{fig:struc}
\end{figure*}%

Our proposed framework extends non-interactive edge-case scenarios by creating an entirely reactive environment in CommonRoad~\cite{Althoff.2017}.
Non-intelligent and deterministic traffic participants (e.g., cars) are substituted by intelligent driving agents,
each attempting to reach a defined target region within a given time interval. 
Every agent is controlled by a sophisticated trajectory planning algorithm, facilitating the execution of complex actions in response to observed actions.
The general structure is depicted in \cref{fig:struc}.
The flexibility in the number of intelligent agents allows for the simulation of non-interactive scenarios with only one agent and interactive scenarios with any number of agents. 
The multi-agent simulation is implemented as a synchronous time-discrete system with planning steps $\Delta t = \SI{0,1}{\second}$.
In each time step, all agents move simultaneously, preventing issues such as non-determinism and guaranteeing that experiments are reproducible~\cite{Bernhard.2020}.
The calculation of multiple agents' actions is split into batches and distributed amongst numerous processes to speed up the computation time. All processes simultaneously calculate the agents' movements and are synchronized after each simulation cycle. The multi-agent framework is written in Python, with some CommonRoad extensions in C++. 

\subsection{Formalization}
\label{ssec:mas}
A simulation $\mathcal{E}:=(S,A)$ is defined by a scenario $S$ and a set of agents $A$.
The scenario $S:=(N, O_s, O_d)$ consists of a street network $N$ built from lanelets, along with a set of potential static and dynamic obstacles, $O_{s}$ and $O_{d}$, respectively.
The state of an agent  $a \in A$ at time step $t$ is defined by the tuple $\mathbf{x}_{a,t} := (x(t), y(t), v(t), \theta(t))$, comprising the position vector $(x(t),y(t))$, the velocity $v(t)$ and the orientation $\theta(t)$.
Each agent $a \in A$ solves a unique planning problem $P_a$, derived from the original benchmark scenarios.
The planning problem $P_a:= (\mathbf{x}_{a,0}, \mathbf{\mathcal{G}}_a, T_{a, max})$ is characterized by the initial state $\mathbf{x}_{a,0}:= (x_0,y_0,v_0,\theta_0)$, and the goal region $\mathbf{\mathcal{G}}_a$ representing the segment of the street network the agent aims to reach within a period of time specific to the agent $T_{a, max}$. 
Specifications for the velocity and orientation within the goal area can also be provided. 
A planning problem $P_a$ is considered successfully solved if the following condition is satisfied:
\begin{equation}
\exists \;t  \geq 0 \; \text{s.t.} \; \mathbf{x}_a(t) \in \mathbf{\mathcal{G}}_a \wedge t \leq T_{a,max}.
\end{equation}
The scenario $S$ symbolizes a global view of the simulation, incorporating all vehicles and agents present up to the simulated time step.
Additionally, for each agent $a \in A$, access is granted solely to a local perspective of the scenario~$S_{a, loc}$, describing the observed environment from the agent's perspective limited to a specific radius around the agent's position.
Every agent is linked to a particular trajectory planner via an interface and computes trajectories in each time step using only the observed environment $S_{a, loc}$.
The stand-alone trajectory planning qualifies each agent to respond dynamically to the observed behavior of the surroundings.
The FRENETIX trajectory planner~\cite{frenetix} is incorporated as a benchmarking trajectory planner and used in the examples presented in~\cref{sec:ne}.
The trajectory planning interface provides the planner with the essential information required for trajectory generation.
In each time step $t$, agents $a \in A$ require information about 
(a) the observable states $ obs_{a}(t)$ of vehicles and obstacles within the visible perimeter of the agent,
\begin{equation}
    obs_{a}(t) = \left\{\mathbf{x}_{i,t} \mid i \in \Tilde{A}\cup \Tilde{O}_{s/d} \right\},
\end{equation}
where $\Tilde{A}\cup \Tilde{O}_{s/d}$ are the other agents and obstacles in the agent's visible field of view regarding the entire scenario~$S_{a, loc}$, and
(b) the motion predictions $pred_a(t)$ of these vehicles for a user-defined prediction horizon $T_{pred}$,
\begin{equation}
pred_a(t) = \left\{\mathbf{x}_{i,t + \tau} \mid \tau \in \left[0, T_{pred}\right],  i \in \Tilde{A} \cup \Tilde{O}_{s/d}\right\}.
\end{equation}
Motion predictions are calculated globally with a prediction horizon of \SI{3}{\second} utilizing trajectory prediction through a neural network~\cite{Geisslinger.2021}. 
Each agent performs a single planning step considering observations and predictions, leading to a new state vector $\mathbf{x}_{a, t+1}$, 
\begin{equation}
\mathbf{x}_{a,t+1} \leftarrow
 f\left(\mathbf{x}_{a,t}, \mathbf{u}_{a,t}, obs_a(t), pred_a(t)\right) \quad \forall a \in A,
\end{equation}
with $\mathbf{u}_{a,t}$ being the input vector containing the dynamically feasible control inputs.
New states $\mathbf{x}_{a, t+1}, \forall a \in A$ are aggregated in the global scenario $S$, where a collision check is performed after each planning iteration~\cite{Pek.2020}. 
Two different types of collisions are detected.
Let $\mathtt{Occ}(\cdot)$ be the occupied space of any object.
An accident is detected when the occupancy $\mathtt{Occ}(\mathbf{x}_{a,t})$ of an agent $a \in A$ at time step $t$ intersects with the occupancy of another agent or obstacle $\mathtt{Occ}(\mathbf{x}_{i,t})$, 
\begin{equation}
    \exists \; i \in A \cup O_{s/d} \;\text{s.t.} \;\mathtt{Occ}(\mathbf{x}_{a,t}) \cap \mathtt{Occ}(\mathbf{x}_{i,t}) \neq \emptyset.
\end{equation}
An accident also occurs when an agent leaves the street network, i.e., if the union of the agent's occupancy and the street network $N$ exceeds the spatial occupancy of the street network itself, 
\begin{equation}
    \mathtt{Occ}(\mathbf{x}_{a,t}) \cup \mathtt{Occ}(N) > \mathtt{Occ}(N).
\end{equation}
Any crashed vehicle is directly removed from the simulation to prevent blocking the roads and to avoid creating different crash scenarios.

\subsection{Benchmarking and Evaluation}
\label{ssec:pse}
Following the completion of the simulation, a retrospective analysis allows for evaluating and comparing the trajectory planning method's efficacy and the criticality of the simulated behavior. 

\subsubsection{Agent Performance Benchmarking}
Performance metrics can be utilized to evaluate and compare the effectiveness of different trajectory planning models.
Based on the simulation outcomes, agents can exhibit the following states contributing to the systematic development of trajectory planning methods:

\begin{itemize}
    \item Goal region reached within the time limit, 
    \item Goal region reached but time limit exceeded,
    \item Time limit exceeded,
    \item Goal region missed,
    \item Trajectory infeasible,
    \item Collision.
\end{itemize}
In the case of a collision, the harm caused by the accident is computed for all vehicles involved to assess the collision's severity \cite{Geisslinger2}.

\subsubsection{Behavioral Safety Benchmarking}

\begin{figure}[!tbp]%
	\centering%
    \includegraphics[width=.95\linewidth, trim=12.5cm 24cm 11cm 18.3cm, clip]{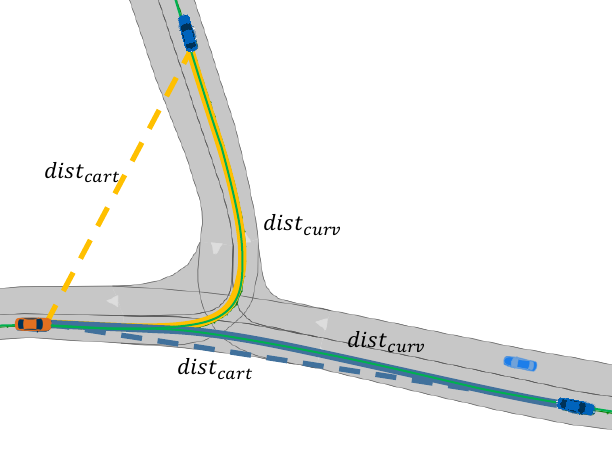}
	\caption{Difference between curvilinear and Cartesian distances. Green: Reference paths of the two curvilinear coordinate systems at the junction spanned for the orange vehicle.}%
	\label{fig:curv}
\end{figure}

Typically, criticality measures rely on distances calculated in Cartesian coordinates~\cite{Katrakazas.2015}.
However, on curved road segments, the curvature significantly distorts the accuracy of Cartesian distances, limiting their relevance in assessing behavioral safety. 
To address this issue, metrics are calculated in curvilinear coordinate systems aligned with feasible driving paths as displayed in \cref{fig:curv}~\cite{Hery.2017}.
In the case of junctions, a coordinate system is established for each theoretically possible driving path to calculate the distances accurately. If metrics can be computed in multiple coordinate systems, we retain the value with higher criticality.
\begin{table}[!b]
\centering
\caption{Metrics for the evaluation of the behavioral safety of AVs}
\renewcommand{\arraystretch}{1.15}
\begin{tabularx}{0.9\linewidth}{l l c}
    \toprule
    \textbf{Acronym} & \textbf{Measure} & \textbf{Source} \\
    \midrule
    HW & Headway & \cite{Jansson.2005} \\
    THW & Time headway  & \cite{Jansson.2005}\\
    TTC & Time-to-collision & \cite{Tamke.2011}\\
    TET & Time-exposed TTC & \cite{Minderhoud.2001}\\
    TIT & Time-integrated TTC & \cite{Westhofen.2023} \\
    TTCE & Time-to-closest-encounter & \cite{Eggert.2014} \\
    DCE & Distance-of-closest-encounter  & \cite{Eggert.2014} \\
    ET & Encroachment time & \cite{Allen.1978} \\
    PET & Post encroachment time & \cite{Allen.1978} \\
    BTN & Break threat number & \cite{Jansson.2005} \\
    STN & Steer threat number  & \cite{Jansson.2005} \\
    PSD & Proportion of minimum stopping distance & \cite{Allen.1978} \\
    MSD & Minimum stopping distance  & \cite{Allen.1978} \\
    \bottomrule
\end{tabularx}
\label{tab:Metrics}
\end{table}
For the optimization of computation time, metrics involving several vehicles are only calculated when the distance between the vehicles is below a threshold value since interactions are only pertinent for vehicles near each other.
Considering oncoming traffic, a frequent occurrence on both rural and urban roads, can lead in highly critical values for an agent each time a car passes. 
To mitigate this, vehicles traveling in the opposite direction are only taken into account if the agent is driving on the oncoming traffic lane, for instance, during overtaking maneuvers. Additionally, this consideration applies if their driving lane intersects with the agent's path such as at an intersection, as illustrated by the semi-transparent blue car in \cref{fig:curv}.
The metrics listed in \cref{tab:Metrics} are incorporated to evaluate the behavioral safety of AVs in the simulated scenario. The precise mathematical formulations are detailed in the respective sources.

\paragraph*{Headway and Time Headway} The headway (HW) at any time step is defined as the curvilinear distance between the front of the agent's vehicle and the rear of a lead vehicle~\cite{Jansson.2005}. The headway is set to infinity in scenarios where no preceding vehicle is identified in the agent's path. By normalizing the headway with the agent's velocity, the time headway (THW) is obtained. THW serves as a criticality measure indicating the time it takes for a trailing vehicle to reach the position of the vehicle ahead~\cite{Jansson.2005}.

\paragraph*{Time-to-Collision and Time-to-Closest-Encounter}
The time-to-collision (TTC) represents the minimum remaining time until two vehicles collide in a car-following scenario or when their trajectories intersect~\cite{Lefevre.2014, Tamke.2011}. 
Assuming constant accelerations, TTC can be computed by considering the HW along with the relative velocity and acceleration between the agent and a leading vehicle~\cite{Hayward.1972, Jansson.2005}.
The time-exposed~TTC~(TET) and the time-integrated~TTC~(TIT) offer insights into behavioral safety at the scenario level upon completion of a simulation~\cite{Minderhoud.2001}. 
TET quantifies the duration the TTC is below a threshold $\tau$. 
To eliminate the dependency of TET on the scenario's duration, it is normalized by dividing it by the total scenario duration.
In a similar vein, the TIT involves integrating the difference between TTC and the threshold $\tau$  over the entire scenario duration, providing a more accurate reflection of criticality compared to TET~\cite{Westhofen.2023}.
The time-to-closest-encounter (TTCE) indicates, at any time step, the remaining time until the distance to surrounding vehicles is minimal~\cite{Eggert.2014}. The minimum distance is referred to as the distance of closest encounter (DCE)~\cite{Eggert.2014}. 
It is evident from the underlying principles that the TTCE is the generalization of TTC to non-collision situations, as TTCE converges towards TTC when DCE approaches zero,

\begin{equation}
    DCE \rightarrow  0 \Leftrightarrow TTCE \rightarrow TTC.
\end{equation}

\paragraph*{Encroachment Time and Post Encroachment Time}
While TTC and its derivatives are effective in car-following situations, their efficacy diminishes at intersections, where complexities arise with vehicles approaching from lateral directions.
To effectively assess these traffic conditions, the encroachment time (ET) and post encroachment time (PET) are used~\cite{Allen.1978}.
The ET is defined as the time required to traverse a conflict area, with the conflict area at intersections delineated as the crossing region of the lanelets.  
On the other hand, PET measures the time until another vehicle enters the conflict area after the agent has crossed it. This metric is frequently utilized in traffic data analysis for estimating traffic density at intersections~\cite{Johnsson.2018, Westhofen.2023}.

\paragraph*{Break Threat Number and Steer Threat Number}
The break threat number (BTN) and steer threat number (STN) are ratios to indicate the difficulty associated with evasive maneuvers. 
They are expressed by the proportion of the necessary longitudinal or lateral acceleration to avoid a collision to the maximum available deceleration or lateral acceleration, respectively~\cite{Jansson.2005}. 
A value $BTN \geq 1$ or $STN \geq 1$ signifies a situation where a collision cannot be avoided through a single longitudinal or lateral maneuver.

\paragraph*{Proportion of Minimum Stopping Distance}
The proportion of minimum stopping distance (PSD) is the quotient between a conflict area and the minimum stopping distance (MSD)~\cite{Allen.1978}. 
The MSD is the distance required under ideal conditions to bring a vehicle to a complete stop.

\section{Results and Analysis}
\label{sec:ne}
Our proposed multi-agent simulation framework is used hereinafter for in-depth scenario and runtime analyses. Initially, we present the outcomes of a runtime analysis conducted with a variable number of intelligent agents, offering insights into the scalability of our framework. 
Subsequently, we illustrate the practical application of the framework by examining scenarios simulated with varying degrees of vehicle interactions.
It is to be mentioned that the results presented herein rely on the motion prediction and trajectory planning method employed. 
As a consequence, outcomes may exhibit variability when integrating alternative approaches. 
All experiments in this evaluation are conducted utilizing the FRENETIX trajectory planner~\cite{frenetix} and WALENET motion prediction~\cite{Geisslinger.2021}, serving as benchmark methods.

\subsection{Runtime Analysis and Computation Performance}
\label{ssec:cp}
A runtime analysis measures the average computation time required for a single planning cycle. 
One planning cycle encompasses the motion prediction, the planning step for each agent, the collision check, and the scenario update.
The analysis is based on four distinct scenarios, each featuring a minimum of 24 vehicles acting as agents.
To cover different scenario complexities, the evaluation encompasses a 6-lane highway, an intersection with 3 to 6 incoming lanes per side, and two recorded scenarios reflecting typical urban German road conditions. 
Each scenario is simulated three times, and the average time across all simulations is calculated.
The assessment is performed on an AMD~Ryzen~9~7950X~processor with 16 cores.
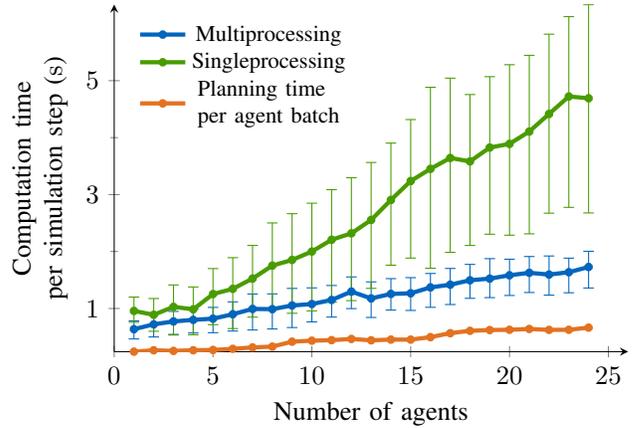
\begin{figure}[t]%
	\centering%
    \input{data/time} 
	\caption{Average computation time per simulation step with single-core and multi-core computation (8 cores). Error bars indicate the first and third quartile of the logged computation time.}
	\label{fig:comp_time}
\end{figure}%
\cref{fig:comp_time} illustrates the average computation time per simulation step with a varying number of intelligent agents.
The simulation is conducted in two configurations: first as a single-core program and second as a multi-core program with up to 8 cores, where batches of agents are distributed among the processes.
When using only one core, the average computation time rises remarkably as the number of agents increases. 
In contrast, distributed computation accelerates the calculation, reducing the average computation time of a single planning cycle by up to \SI{74}{\percent} for 24 agents.
The distribution of the computation time varies significantly but consistently remains smaller for the multi-core approach.
The average calculation time per batch for generating new trajectories for each agent reveals that the planning time per agent remains constant when multi-core computation is used.
However, the experiments with 9 and 17 agents exhibit a gradual rise in the calculation time, indicating that the number of agents per batch has grown.

\subsection{Interactive Behavior of Multiple Agents}
\label{ssec:ib}
We demonstrate the multi-agent simulation framework's use case by qualitatively comparing vehicle interaction types. 
A lane merging scenario and an intersection situation are solved with two agents per scenario. Our ego vehicle (orange car in \cref{fig:diff1}ff) is the intelligent AV agent, while the traffic participant (green in \cref{fig:diff1}ff) varies its type of interaction in three cases: (a) full deterministic with non-interaction, (b)~semi-deterministic with variable velocity, (c) fully intelligent with trajectory planner and full interaction.
\subsubsection{Lane Merging Scenario}

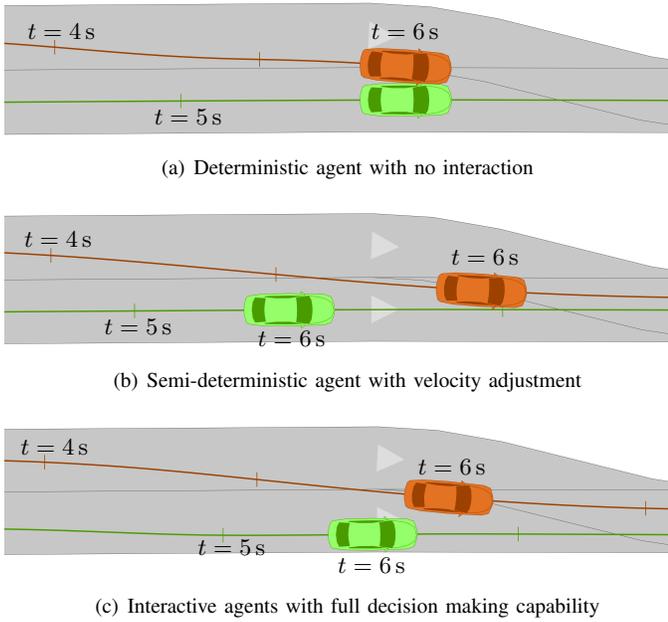
\begin{figure}[!t]%
\centering%
\subfigure[Deterministic agent  with no interaction]{
\centering
\input{figures/merge1}
    \label{fig:1ag}
    }
\subfigure[Semi-deterministic agent with velocity adjustment]{
\centering
\input{figures/merge2}
    \label{fig:2idm}
    }
\subfigure[Interactive agents with full decision making capability]{
\centering
\input{figures/merge3}
    \label{fig:2ag}
    }
 \caption{Lane merging scenario: Comparison of the driving interactions. The orange vehicle is an intelligent agent, while the interaction abilities of the green car are altered in each simulation.}%
	\label{fig:diff1}
\end{figure}%

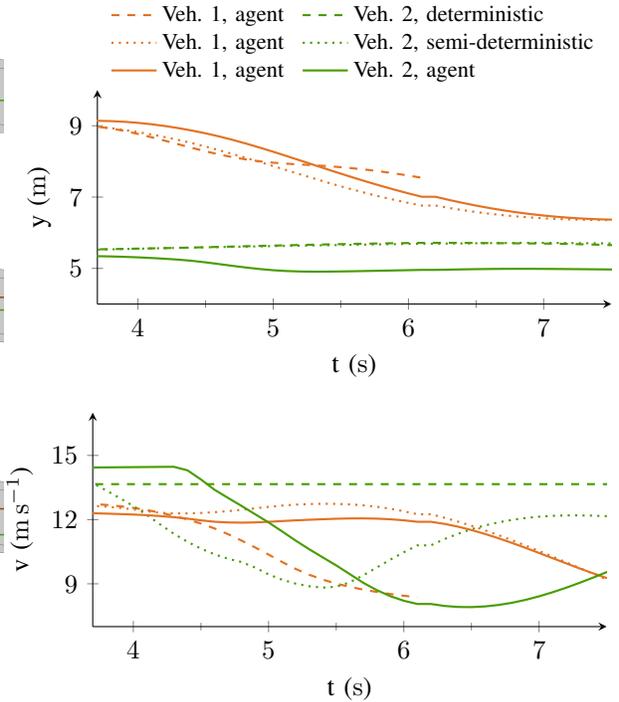
\begin{figure}[!t]%
	\centering%
 \subfigure{
 \centering
         \input{data/diff} 
 }
   \subfigure{
  \centering
         \input{data/diff2}
 \label{fig:diff_plot12}
 }
	\caption{Lane merging scenario: Comparison of trajectory and velocity profiles.}%
	\label{fig:diff_plot1}
\end{figure}

In \cref{fig:diff1}, the same time step is illustrated for all three interaction variations of the orange car. 
The differences in the trajectories are highlighted in \cref{fig:diff_plot1}, emphasizing the adjustments in the lateral vehicle positions on the lanelet and velocities.
The orange vehicle, representing an intelligent agent, seeks to merge into the lane occupied by the green car, which possesses altering abilities.
In \cref{fig:1ag}, the green car adheres to a recorded trajectory with a constant velocity, resulting in a collision at the depicted time step.
When the green vehicle adjusts only the velocity, as in \cref{fig:2idm}, it decelerates to evade a collision, leaving a gap for merging.
In the final setting, where both agents can comprehensively react to the observed behavior, an additional lateral maneuver is apparent, with the green vehicle swerving to the right.
In all three simulations, the orange agent maintains a similar trajectory. 
However, when the green agent can interact, the orange vehicle refrains from initiating a braking maneuver to avoid the collision, as observed in the velocity profile of the deterministic scenario. 
If the green vehicle can reduce its velocity, it increases the spacing to allow the orange one to retract. 
This results in the highest DCE of all three simulations (\cref{tab:met_merge}). Subsequently, the vehicle accelerates again due to the large gap,  as seen in \cref{fig:diff_plot1}, yielding the smallest TTC of the non-colliding simulations.
In the interactive simulation,  the evasive maneuver of the green vehicle starts later, and an additional side-wise movement is observed, resulting in a smaller DCE. However, given the lower velocity during the lane merge, the TTC is four times as high at \SI{2.3}{\second}, making the situation, in summary, less critical.  
\begin{table}[!h]%
    \centering
    \renewcommand{\arraystretch}{1.15}
    \caption{Criticality metrics for the lane merging scenario}
    \begin{tabularx}{0.98\linewidth}{l c c c c}
        \toprule
        \makecell[l]{\textbf{Degree of}\\ \textbf{Interaction}} & \makecell{\textbf{min. DCE}\\ (\si{\metre})} &\makecell{\textbf{min. TTC} \\(\si{\second}) } &\makecell{\textbf{max. BTN} \\ (-) } & \makecell{\textbf{Agent} \\\textbf{collided}}\\
        \midrule
       No Interaction  & 0 & 0 & 0.5 & True\\
        Velocity adjusting &  1.9 & 0.45 &  0.43 & False\\
       Interactive & 0.5 & 2.3 &  0.44 & False\\
       \bottomrule
    \end{tabularx}
    \label{tab:met_merge}
\end{table}

\subsubsection{Intersection Scenario}
\begin{figure*}[!tbp]%
\centering%
 \subfigure[Deterministic agent  with no interaction]{
\centering
 \input{figures/tjunc1}
    \label{fig:1ag_2}
}
\hfill
 \subfigure[Semi-deterministic agent with velocity adjustment]{
\centering
\input{figures/tjunc2}
    \label{fig:2idm_2}
}
\hfill
 \subfigure[Interactive agents with full decision making capability]{
\centering
\input{figures/tjunc3}
    \label{fig:2ag_2}
}
	\caption{Intersection scenario: Comparison of the driving interactions. The orange vehicle is an intelligent agent, while the interaction abilities of the green car are altered in each simulation.}%
	\label{fig:diff2}
\end{figure*}
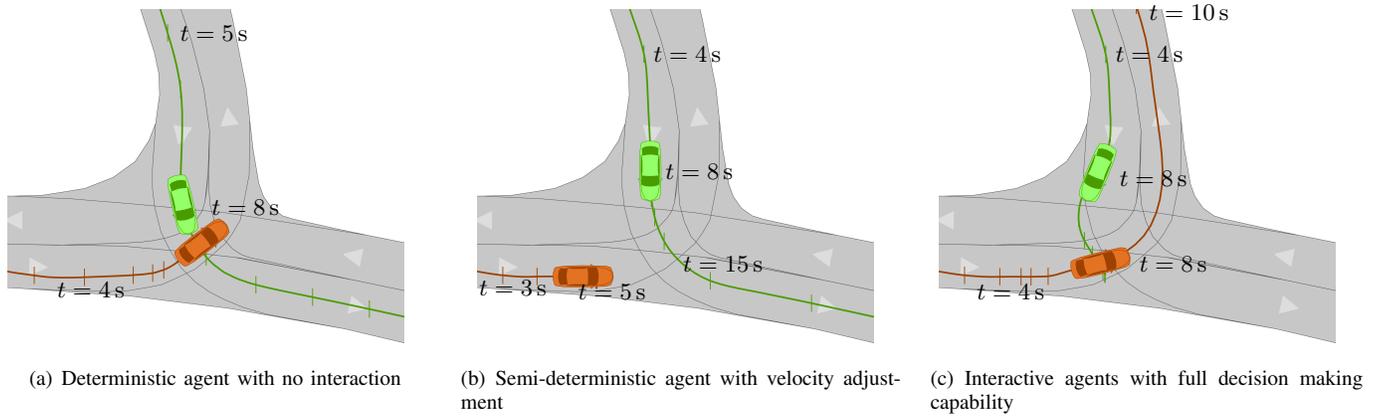
\begin{figure}[!htbp]%
	\centering%
         \input{data/diff_tjunc2}
	\caption{Comparison of the velocity profile in the recorded and interactive intersection scenario.}%
	\label{fig:diff_plot2}
\end{figure}
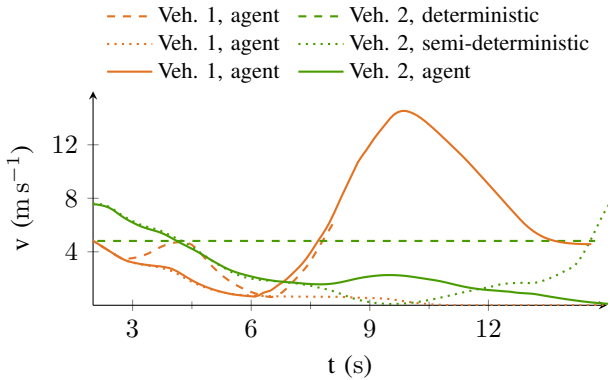

%
Simulating an intersection with the three described settings reveals markedly divergent behaviors.
With recorded traffic, a collision occurs in \cref{fig:1ag_2}, while depending on the behavior of the green vehicle, the agent in orange reacts in a contrasting manner in \cref{fig:2idm_2} and \cref{fig:2ag_2}. 
When only the velocity of the green agent is altered, the orange vehicle comes to a halt, allowing the green car to pass, thereby failing to traverse the intersection within the given time limit. 
Nevertheless, the green agent proceeds through the intersection at a lower velocity than the recorded trajectory (\cref{fig:diff_plot2}), resulting in an ET value of \SI{30}{\second}. 
In the fully interactive simulation (\cref{fig:2ag_2}), both vehicles decelerate, with the green one attempting to avoid a potential collision by driving an arc-like trajectory. 
Consequently, the orange agent accelerates and forces through the intersection in front of the green vehicle. 
The minimum distance is \SI{1}{\metre} less than in the previous case. Also, the ET and PET values are significantly smaller, indicating a more critical situation than in \cref{fig:2idm_2}.

\begin{table}[!htbp]%
    \centering
    \renewcommand{\arraystretch}{1.15}
    \caption{Criticality Metrics for the intersection scenario}
    \begin{tabularx}{0.85\linewidth}{l c c c c} % c|c|}
        \toprule
        \makecell[l]{\textbf{Degree of}\\ \textbf{Interaction}} & \makecell{\textbf{min. DCE} \\ (\si{\metre})} &  \makecell{\textbf{ET}\\ (\si{\second})}& \makecell{\textbf{ PET}\\ (\si{\second})}&  \makecell{\textbf{Agent} \\\textbf{collided} }\\
        \midrule
       Recorded  & 0 &% 1.2 & 1.3 &
       n.a. & 0 & True\\
        Velocity adjusting  & 2.6 & %0.5  & 0.001&
       30 & inf & False\\
       Interaction &  1.8 &%  0.5 & 0.14 &
       11& 9 & False\\
       \bottomrule
    \end{tabularx}
    
    \label{tab:met_tjunc}
\end{table}

\section{Discussion}
Our presented multi-agent simulation framework has the possibility to simulate autonomous vehicles in fully interactive scenarios with a variable number of intelligent agents.
In \cref{ssec:cp}, we analyzed the runtime of our framework. Irrespective of the number of agents, the framework's computation time per time step exceeds the simulated discretized time step size of \SI{0.1}{\second}. A larger number of agents leads to an observable increase in computation time.
Consequently, achieving real-time simulation is unattainable even with a planning method capable of real-time calculation.
This limitation stems from the simulation completing all agents' planning steps before advancing to the next time step.
While hindering real-time performance, this synchronous nature of the simulation framework ensures determinism and reproducibility of experiments.

In particular, the multi-agent framework offers an advantage when investigating the planning results of agents in diverse situations. In a single simulation run, multiple agents can plan their trajectories, expediting experiments by eliminating the need to initiate a separate simulation for each agent.
The analysis of the interactive behavior of two agents in \cref{ssec:ib} illustrates the capability of our framework to simulate complex driving interactions while ensuring a physically feasible driving behavior of all vehicles. 
Currently, the simulation requires every agent to be controlled by the same planning method, limiting the diversity of observed behaviors and potentially differing from representing real-world situations. 
However, the existing outcomes underscore the importance of interactive driving simulations. 
They are crucial for developing robust trajectory planning methods, as even minor modifications in the environmental behavior can lead to substantial differences in the planned trajectories. 

\section{Conclusion \& Outlook}
This paper introduces a multi-agent simulation framework designed to simulate non-cooperative interactive driving for AVs in complex environments. 
The framework serves as an instrument for testing and evaluating different trajectory planning methods in environments where multiple agents simultaneously navigate to individual destinations.
The findings highlight the impact of vehicle interactions on AV trajectory planning algorithms; therefore, the framework could also be used for V2V research.
Future studies should explore cooperative behavior and inter-vehicle communication to enhance autonomous driving. Additionally, experiment variability could be increased by integrating various planning approaches into the simulation framework.
Training trajectory planning methods using reinforcement learning within the multi-agent framework could also greatly benefit the research community.

\bibliographystyle{IEEEtran}
\bibliography{refs}

\end{document}

%% file: data/time.tex
% \documentclass{article}
% %
% %% Einige Packages, die nützlich sind für die Erstellung von Grafiken:
% \usepackage[utf8]{inputenc}
% \usepackage{soulutf8}
% \usepackage{pgfplots}
% \usetikzlibrary{plotmarks}
% \usepgfplotslibrary{groupplots}
% \usepgfplotslibrary{statistics}
% \usepgfplotslibrary{external}
% % \tikzexternalize
% % Für das Einbinden von Tabellen in Bilder können diese Packages hilfreich sein:
% %\usepackage{harveyBalls}
% %\usepackage{fontawesome}
% %\usepackage{multirow}
% \usepackage[per-mode=symbol]{siunitx-v2}
% \begin{document}
% 	\definecolor{TUMGreen}{RGB}{162, 173,0}
% 	\definecolor{TUMBlue}{RGB}{0,101,189}
% 	\definecolor{TUMOrange}{RGB}{227,114,34}

\begin{tikzpicture}%[trim axis left, ,trim axis right]

	\begin{axis}[
		boxplot/draw direction=y,
		width=0.95\linewidth,
		height=0.7\linewidth,
        xmin = 0, xmax = 26, 
        legend style={cells={align=left}},
%		symbolic x coords={Acceleration},
		% xtick={1,2,3},
		% ytick={0,2,4,6},
		% xticklabels = {Aggressive, Defensive, Normal },
        ytick = {1,3,5},
		minor y tick num=1,
		axis x line = bottom,
		axis y line = left,
        ylabel style={align=center},
		ylabel={Computation time \\ per simulation step (\si[per-mode=symbol]{\second})},
		xlabel={Number of agents},
   legend pos=north west,
   	legend style= {nodes={scale=0.85, transform shape},
		},
  legend style={draw=none},
 %   y label style={yshift=-1em},
 %   legend cell align = {left},
		%		ybar interval=0.5,
		%	legend to name = leg,
		%	legend columns = 3,
		%	legend style={
			%		legend cell align = left,
			%		%		legend pos=outer north east},
		%		at = {(1.05, 0.5)},
		%		anchor = west},
	]

	% \addplot+ [mark=*, mark options={scale=0.3, fill, TUMOrange}, fill=TUMOrange!30,draw=TUMOrange,boxplot={draw position=1, box extend = 0.5}] table[y=a, col sep=comma] {figures/IRL_eval/RMSE_Agg_training_set.csv};
	% \addplot+ [mark=*, mark options={scale=0.3, fill, TUMBlue}, fill=TUMBlue!30,draw=TUMBlue,boxplot={draw position=2, box extend = 0.}] table[col sep=comma] {data/single.csv};

	\addplot+ [ultra thick,mark=*, mark options={scale=0.4, fill, TUMBlue},draw=TUMBlue,
        error bars/.cd, 
        y explicit,
        y dir=both,
         error bar style={color=TUMBlue},
 ] table[x = x, y = Mean_Multi, y error plus = dquart3_multi, y error minus = dquart1_multi,col sep=semicolon] {data/times_neu.csv};

 % 	\addplot+ [mark=*, mark options={scale=0.3, fill, TUMBlue},draw=TUMBlue,
 %        error bars/.cd, 
 %        y explicit,
 %        y dir=both,
 % ] table[x = x, y = Mean_Multi, y error plus = dmax_multi, y error minus = dmin_multi,col sep=semicolon] {data/times.csv};
 \addlegendentry{Multiprocessing} % with \\ at most 8 processes}
 
    \addplot+ [ultra thick,mark=*, mark options={scale=0.4, fill, TUMGreen}, draw=TUMGreen,
        error bars/.cd, 
        y explicit,
        y dir=both,
        error bar style={color=TUMGreen},
        ] table[x = x, y = Mean_single, y error plus = dquart3_single, y error minus = dquart1_single, col sep=semicolon] {data/times_neu.csv};
     \addlegendentry{Singleprocessing}

         \addplot+ [ultra thick, mark=*, mark options={scale=0.4, fill, TUMOrange}, draw=TUMOrange,
        % error bars/.cd, 
        % y explicit,
        % y dir=both,
        % error bar style={color=TUMGreen},
        ] table[x = x, y = sim_step, col sep=semicolon] {data/times_neu.csv};
     \addlegendentry{Planning time\\ per agent batch}

    % \addplot+ [mark=*, mark options={scale=0.3, fill, TUMGreen}, draw=TUMGreen,
    %     error bars/.cd, 
    %     y explicit,
    %     y dir=both,] table[x = x, y = Mean_single, y error plus = dmax_single, y error minus = dmin_single, col sep=semicolon] {data/times.csv};
\end{axis}

\end{tikzpicture}

% \end{document}

%% file: figures/merge1.tex
% \documentclass{article}
% %
% %% Einige Packages, die nützlich sind für die Erstellung von Grafiken:
% \usepackage[utf8]{inputenc}
% \usepackage{soulutf8}
% \usepackage{pgfplots}
% \usepackage{siunitx}
% \usetikzlibrary{plotmarks}
% \usepgfplotslibrary{groupplots}
% \usepgfplotslibrary{statistics}
% \usepgfplotslibrary{external}
% % \tikzexternalize
% % Für das Einbinden von Tabellen in Bilder können diese Packages hilfreich sein:
% %\usepackage{harveyBalls}
% %\usepackage{fontawesome}
% %\usepackage{multirow}
% % \usepackage[per-mode=symbol]{siunitx-v2}
% \begin{document}
% 	\definecolor{TUMGreen}{RGB}{162, 173,0}
% 	\definecolor{TUMBlue}{RGB}{0,101,189}
% 	\definecolor{TUMOrange}{RGB}{227,114,34}

\begin{tikzpicture}%[trim axis left, ,trim axis right]

\node[anchor=south west,inner sep=0] (Bild) at (0,0) {\includegraphics[width=\linewidth]{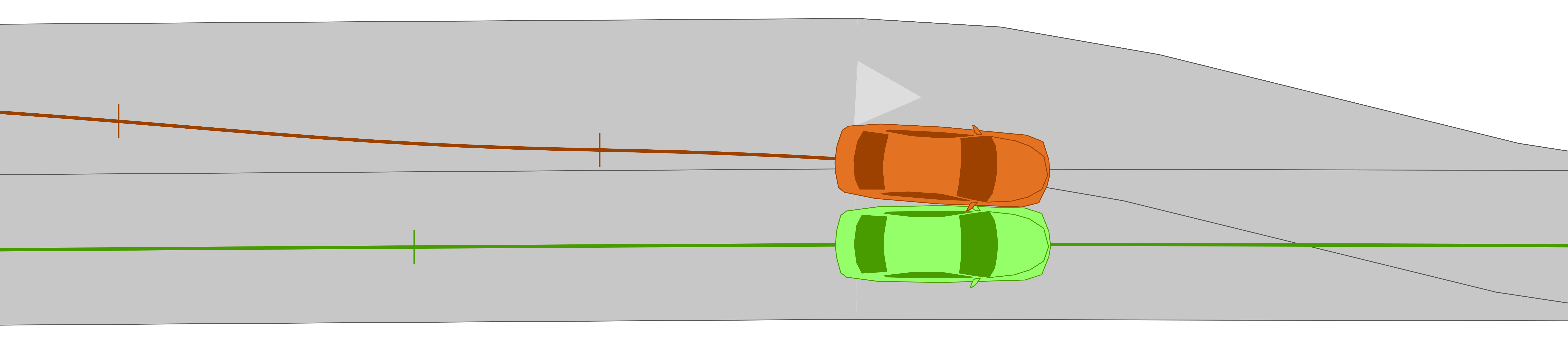}};
\begin{scope}[x=(Bild.south east),y=(Bild.north west)]
     \node[font = {\small}] at (0.275,0.18){ $t= \SI{5}{\second}$};
     \node[font = {\small}] at (0.6,0.76){ $t= \SI{6}{\second}$};
     \node[font = {\small}] at (0.085,0.76){ $t= \SI{4}{\second}$};
\end{scope}

% \node[inner sep=0pt] at (0,0)
%     {\includegraphics[width=\textwidth]{figures/1agents.png}};
% \node[] at (-1,-0.5){$t= \SI{5}{\second}$};
\end{tikzpicture}

% \end{document}

%% file: figures/merge2.tex
% \documentclass{article}
% %
% %% Einige Packages, die nützlich sind für die Erstellung von Grafiken:
% \usepackage[utf8]{inputenc}
% \usepackage{soulutf8}
% \usepackage{pgfplots}
% \usepackage{siunitx}
% \usetikzlibrary{plotmarks}
% \usepgfplotslibrary{groupplots}
% \usepgfplotslibrary{statistics}
% \usepgfplotslibrary{external}
% % \tikzexternalize
% % Für das Einbinden von Tabellen in Bilder können diese Packages hilfreich sein:
% %\usepackage{harveyBalls}
% %\usepackage{fontawesome}
% %\usepackage{multirow}
% % \usepackage[per-mode=symbol]{siunitx-v2}
% \begin{document}
% 	\definecolor{TUMGreen}{RGB}{162, 173,0}
% 	\definecolor{TUMBlue}{RGB}{0,101,189}
% 	\definecolor{TUMOrange}{RGB}{227,114,34}

\begin{tikzpicture}%[trim axis left, ,trim axis right]

\node[anchor=south west,inner sep=0] (Bild) at (0,0) {\includegraphics[width=\linewidth]{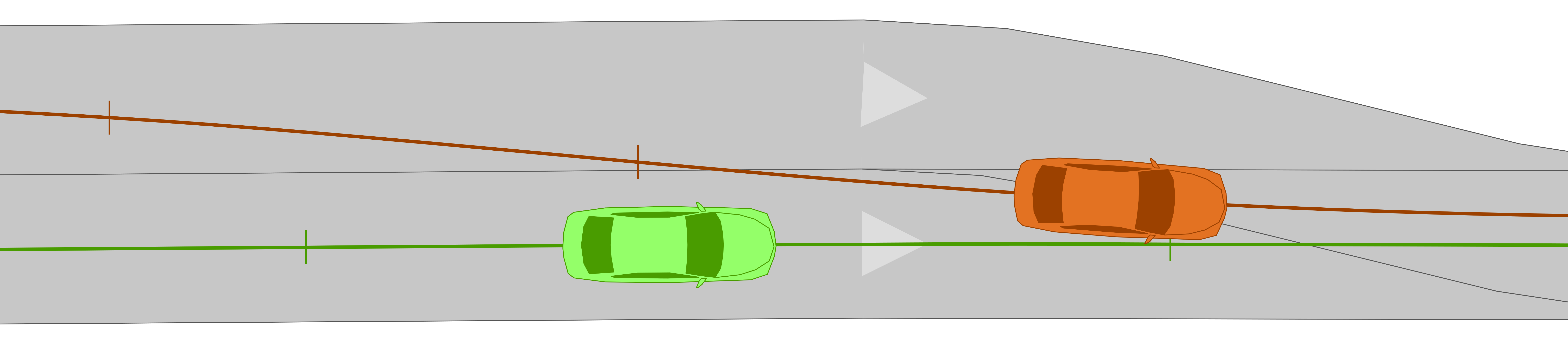}};
\begin{scope}[x=(Bild.south east),y=(Bild.north west)]
     \node[font = {\small}] at (0.2,0.18){$t= \SI{5}{\second}$};
     \node[font = {\small}] at (0.43,0.1){$t= \SI{6}{\second}$};
     % \node[font = {\small}] at (0.755,0.18){$t= \SI{7}{\second}$};
     \node[font = {\small}] at (0.08,0.77){$t= \SI{4}{\second}$};
     \node[font = {\small}] at (0.72,0.65){$t= \SI{6}{\second}$};     
\end{scope}

% \node[inner sep=0pt] at (0,0)
%     {\includegraphics[width=\textwidth]{figures/1agents.png}};
% \node[] at (-1,-0.5){$t= \SI{5}{\second}$};
\end{tikzpicture}

% \end{document}

%% file: figures/merge3.tex
% \documentclass{article}
% %
% %% Einige Packages, die nützlich sind für die Erstellung von Grafiken:
% \usepackage[utf8]{inputenc}
% \usepackage{soulutf8}
% \usepackage{pgfplots}
% \usepackage{siunitx}
% \usetikzlibrary{plotmarks}
% \usepgfplotslibrary{groupplots}
% \usepgfplotslibrary{statistics}
% \usepgfplotslibrary{external}
% % \tikzexternalize
% % Für das Einbinden von Tabellen in Bilder können diese Packages hilfreich sein:
% %\usepackage{harveyBalls}
% %\usepackage{fontawesome}
% %\usepackage{multirow}
% % \usepackage[per-mode=symbol]{siunitx-v2}
% \begin{document}
% 	\definecolor{TUMGreen}{RGB}{162, 173,0}
% 	\definecolor{TUMBlue}{RGB}{0,101,189}
% 	\definecolor{TUMOrange}{RGB}{227,114,34}

\begin{tikzpicture}%[trim axis left, ,trim axis right]

\node[anchor=south west,inner sep=0] (Bild) at (0,0) {\includegraphics[width=\linewidth]{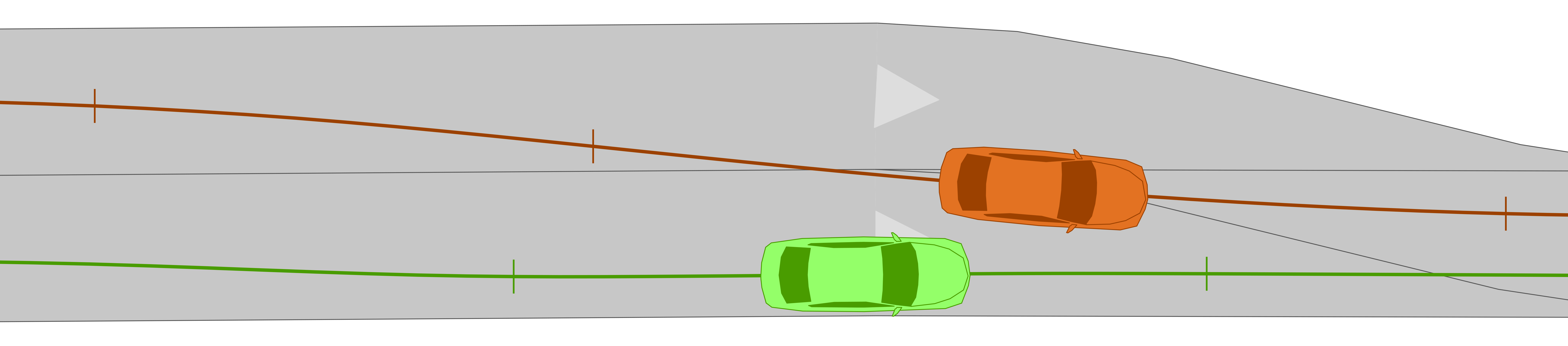}};
\begin{scope}[x=(Bild.south east),y=(Bild.north west)]
     \node[font = {\small}] at (0.34,0.12){$t= \SI{5}{\second}$};
      \node[font = {\small}] at (0.55,0.0){$t= \SI{6}{\second}$};
     % \node[font = {\small}] at (0.78,0.12){$t= \SI{7}{\second}$};
     \node[font = {\small}] at (0.0745,0.8){$t= \SI{4}{\second}$};
     % \node[font = {\small}] at (0.39,0.68){$t= \SI{5}{\second}$};
     \node[font = {\small}] at (0.67,0.66){$t= \SI{6}{\second}$};     
     % \node[font = {\small}] at (0.97,0.5){$t= \SI{7}{\second}$};
     
\end{scope}

% \node[inner sep=0pt] at (0,0)
%     {\includegraphics[width=\textwidth]{figures/1agents.png}};
% \node[] at (-1,-0.5){$t= \SI{5}{\second}$};
\end{tikzpicture}

% \end{document}

%% file: data/diff.tex
%%%%over t
\begin{tikzpicture}[trim axis left, ,trim axis right]

	\begin{axis}[
		boxplot/draw direction=y,
		width=.95\linewidth,
		height=0.5\linewidth,
        xmin = 3.7, xmax = 7.5,
        ymin = 4, ymax = 10,  
        % legend style={cells={align=left}},
%		symbolic x coords={Acceleration},
		xtick={4.0,5.0,6.0,7.0},
		ytick={5,7,9},
		minor x tick num=1,
        % minor y tick num=1,
		axis x line = bottom,
		axis y line = left,
        ylabel style={align=center},
		ylabel={y (\si[per-mode=symbol]{\metre})},
		xlabel={t (\si[per-mode=symbol]{\second})},
%    % legend pos=north east,
 legend style={draw=none},
 legend columns=2, 
	legend style= {at={(0.5,1)},anchor=south, nodes={scale=0.85, transform shape},
		legend cell align = left, /tikz/column 2/.style={
                column sep=5pt}},
	]

    \addplot [thick, draw=TUMOrange,dashed,] 
            table[x = t, y = 1_0_y, col sep=comma] {data/diff.csv};
    \addlegendentry{Veh. 1, agent}
	\addplot [thick, draw=TUMGreen,dashed ] 
            table[x = t, y = 2_0_y, col sep=comma] {data/diff.csv};
            \addlegendentry{Veh. 2, deterministic}
            
    \addplot [thick, draw=TUMOrange, dotted ] 
            table[x = t, y = 1_idm_y, col sep=comma] {data/diff.csv};
            \addlegendentry{Veh. 1, agent}
	\addplot [thick, draw=TUMGreen, dotted] 
            table[x = t, y = 2_idm_y, col sep=comma] {data/diff.csv};
            \addlegendentry{Veh. 2, semi-deterministic}

    \addplot [thick, draw=TUMOrange, ] 
            table[x = t, y = 1_ag_y, col sep=comma] {data/diff.csv};
            \addlegendentry{Veh. 1, agent}
	\addplot [thick, draw=TUMGreen ] 
            table[x = t, y = 2_ag_y, col sep=comma] {data/diff.csv};
            \addlegendentry{Veh. 2, agent}

\end{axis}

\end{tikzpicture}

%% file: data/diff2.tex
%%%% over t
\begin{tikzpicture}[trim axis left, ,trim axis right]

	\begin{axis}[
		boxplot/draw direction=y,
		width=.95\linewidth,
		height=0.5\linewidth,
        xmin = 3.7, xmax = 7.5,
        % legend style={cells={align=left}},
%		symbolic x coords={Acceleration},
		xtick={4.0,5.0,6.0,7.0},
        ymin = 7, ymax = 17,        
        legend style={cells={align=left}},
%		symbolic x coords={Acceleration},
		ytick={9,12,15},
		% xticklabels = {Aggressive, Defensive, Normal },
		% minor y tick num=1,
		axis x line = bottom,
		axis y line = left,
  minor x tick num=1,
        % minor y tick num=1,
		%		ybar = 0.05cm,
		%		bar width =  0.5pt,
%		xmax = 3.5,
        ylabel style={align=center},
		ylabel={v (\si[per-mode=reciprocal]{\metre\per\second})},
		xlabel={t (\si[per-mode=reciprocal]{\second})},
   legend pos=north west,
 legend style={draw=none},
	]

    \addplot [thick, draw=TUMOrange,dashed,] 
            table[x = t, y = 1_0_v, col sep=comma] {data/diff.csv};
    % \addlegendentry{Veh. 1, agent}
	\addplot [thick, draw=TUMGreen,dashed ] 
            table[x = t, y = 2_0_v, col sep=comma] {data/diff.csv};
            % \addlegendentry{Veh. 2, non-reactive}
            
    \addplot [thick, draw=TUMOrange, dotted ] 
            table[x = t, y = 1_idm_v, col sep=comma] {data/diff.csv};
            % \addlegendentry{Veh. 1, agent}
	\addplot [thick, draw=TUMGreen, dotted] 
            table[x = t, y = 2_idm_v, col sep=comma] {data/diff.csv};
            % \addlegendentry{Veh. 2, longitudinal reactions only}

    \addplot [thick, draw=TUMOrange, ] 
            table[x = t, y = 1_ag_v, col sep=comma] {data/diff.csv};
            % \addlegendentry{Veh. 1, agent}
	\addplot [thick, draw=TUMGreen ] 
            table[x = t, y = 2_ag_v, col sep=comma] {data/diff.csv};
            % \addlegendentry{Veh. 2, agent}

\end{axis}

\end{tikzpicture}

%% file: figures/tjunc1.tex
% \documentclass{article}
% %
% %% Einige Packages, die nützlich sind für die Erstellung von Grafiken:
% \usepackage[utf8]{inputenc}
% \usepackage{soulutf8}
% \usepackage{pgfplots}
% \usepackage{siunitx}
% \usetikzlibrary{plotmarks}
% \usepgfplotslibrary{groupplots}
% \usepgfplotslibrary{statistics}
% \usepgfplotslibrary{external}
% % \tikzexternalize
% % Für das Einbinden von Tabellen in Bilder können diese Packages hilfreich sein:
% %\usepackage{harveyBalls}
% %\usepackage{fontawesome}
% %\usepackage{multirow}
% % \usepackage[per-mode=symbol]{siunitx-v2}
% \begin{document}
% 	\definecolor{TUMGreen}{RGB}{162, 173,0}
% 	\definecolor{TUMBlue}{RGB}{0,101,189}
% 	\definecolor{TUMOrange}{RGB}{227,114,34}

\begin{tikzpicture}%[trim axis left, ,trim axis right]

\node[anchor=south west,inner sep=0] (Bild) at (0,0) {\includegraphics[width=0.29\linewidth]{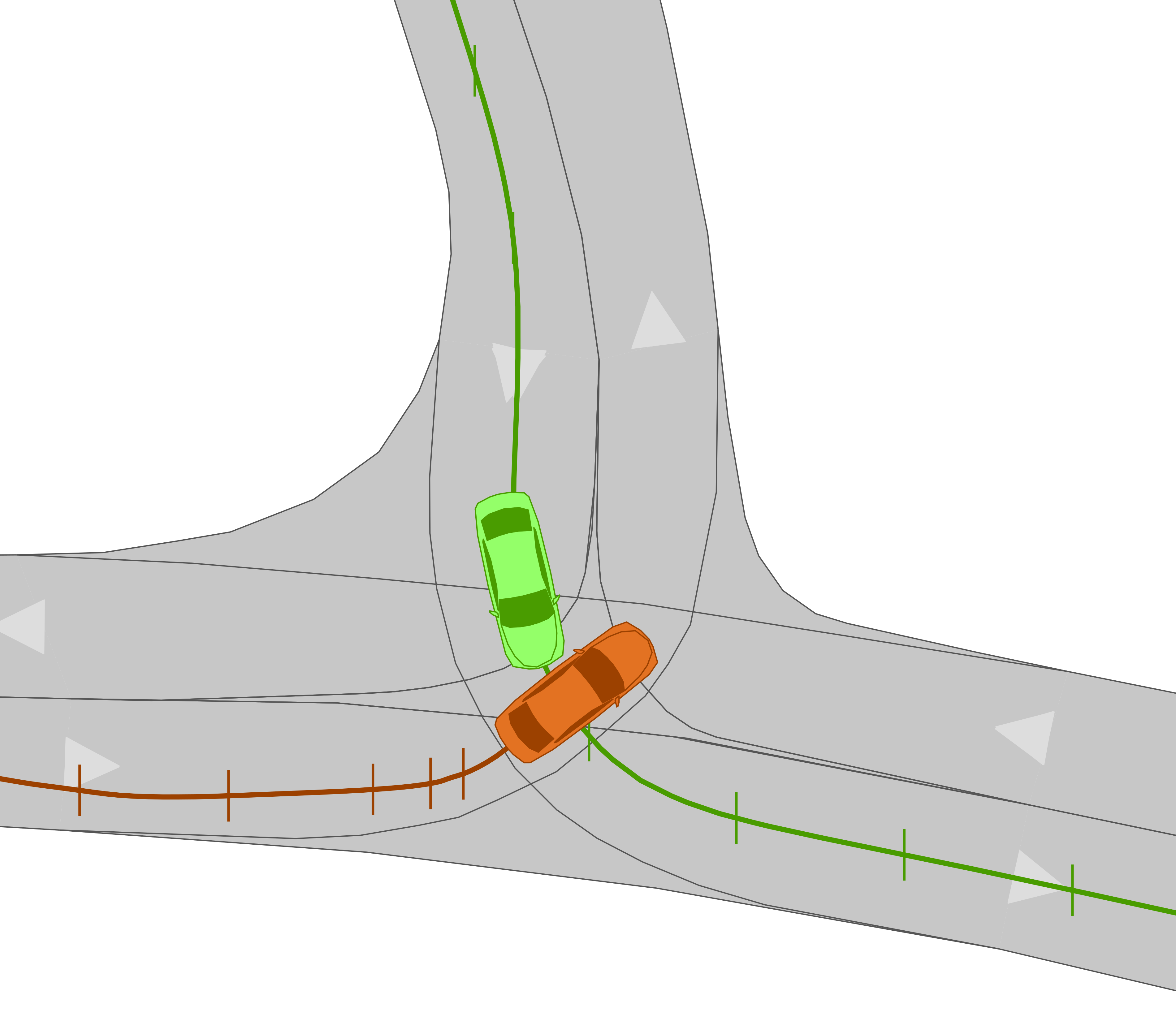}};
\begin{scope}[x=(Bild.south east),y=(Bild.north west)]
     \node[font = {\small}] at (0.52,0.93){$t= \SI{5}{\second}$};
     % \node[font = {\small}] at (0.55,0.765){$t= \SI{6}{\second}$};
     % \node[font = {\small}] at (0.55,0.6){$t= \SI{7}{\second}$};
    \node[font = {\small}] at (0.6,0.425){$t= \SI{8}{\second}$};
     % \node[font = {\small}] at (0.418,0.2){$t= \SI{7}{\second}$};
     % \node[font = {\small}] at (0.325,0.19){$t= \SI{5}{\second}$};
     \node[font = {\small}] at (0.21,0.185){$t= \SI{4}{\second}$};
     % \node[font = {\small}] at (0.09,0.19){$t= \SI{3}{\second}$};
\end{scope}

% \node[inner sep=0pt] at (0,0)
%     {\includegraphics[width=\textwidth]{figures/1agents.png}};
% \node[] at (-1,-0.5){$t= \SI{5}{\second}$};
\end{tikzpicture}

% \end{document}

%% file: figures/tjunc2.tex
% \documentclass{article}
% %
% %% Einige Packages, die nützlich sind für die Erstellung von Grafiken:
% \usepackage[utf8]{inputenc}
% \usepackage{soulutf8}
% \usepackage{pgfplots}
% \usepackage{siunitx}
% \usetikzlibrary{plotmarks}
% \usepgfplotslibrary{groupplots}
% \usepgfplotslibrary{statistics}
% \usepgfplotslibrary{external}
% % \tikzexternalize
% % Für das Einbinden von Tabellen in Bilder können diese Packages hilfreich sein:
% %\usepackage{harveyBalls}
% %\usepackage{fontawesome}
% %\usepackage{multirow}
% % \usepackage[per-mode=symbol]{siunitx-v2}
% \begin{document}
% 	\definecolor{TUMGreen}{RGB}{162, 173,0}
% 	\definecolor{TUMBlue}{RGB}{0,101,189}
% 	\definecolor{TUMOrange}{RGB}{227,114,34}

\begin{tikzpicture}%[trim axis left, ,trim axis right]

\node[anchor=south west,inner sep=0] (Bild) at (0,0) {\includegraphics[width=0.29\linewidth]{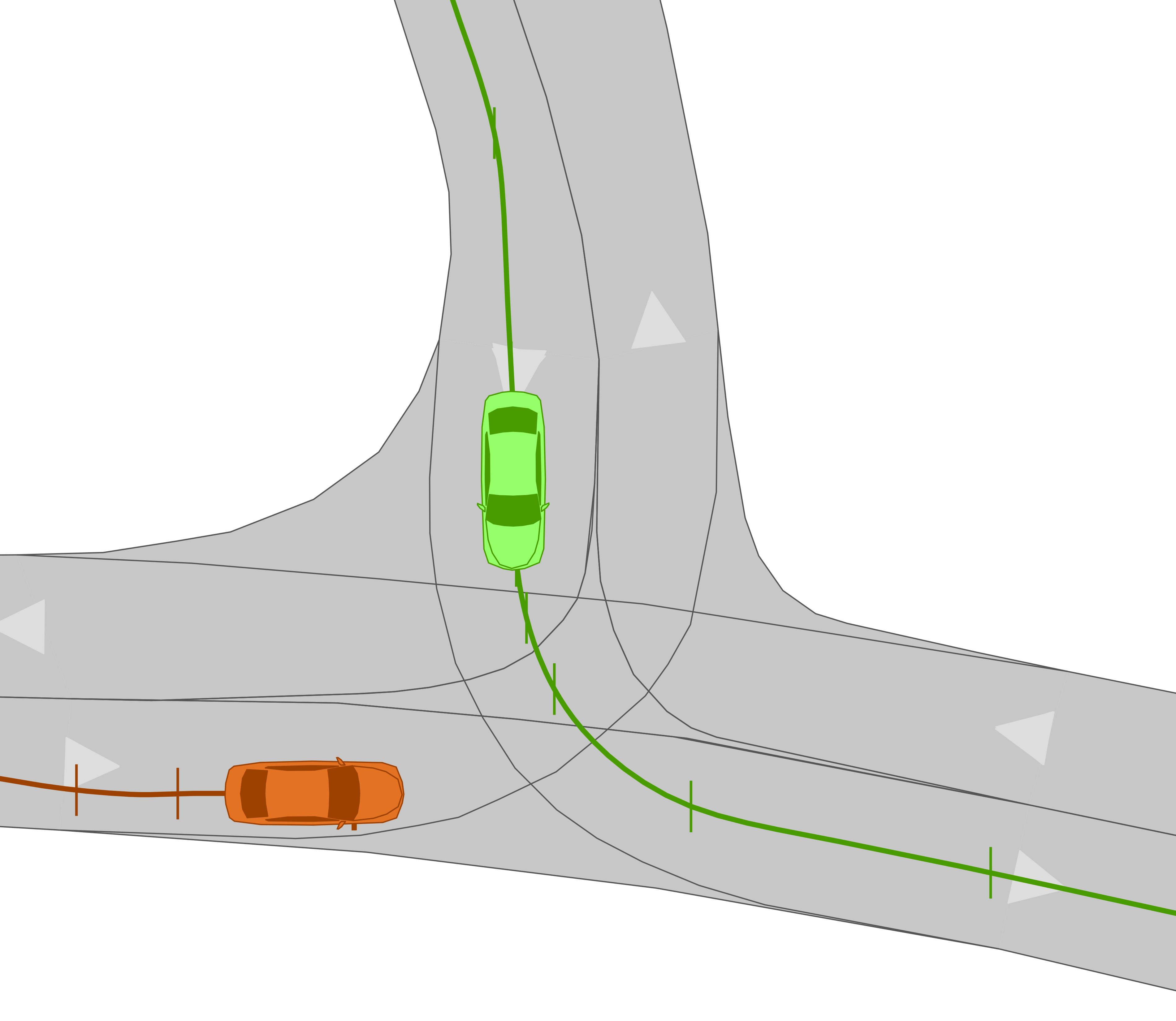}};
\begin{scope}[x=(Bild.south east),y=(Bild.north west)]
     \node[font = {\small}] at (0.53,0.87){$t= \SI{4}{\second}$};
     % \node[font = {\small}] at (0.485,0.72){$t= \SI{5}{\second}$};
     % \node[font = {\small}] at (0.49,0.66){$t= \SI{6}{\second}$};
     
% \node[font = {\small}] at (0.55,0.625){$t= \SI{7}{\second}$};
\node[font = {\small}] at (0.56,0.53){$t= \SI{8}{\second}$};

    % \node[font = {\small}] at (0.57,0.44){$t= \SI{12}{\second}$};
    % \node[font = {\small}] at (0.51,0.39){$t= \SI{13}{\second}$};
    % \node[font = {\small}] at (0.535,0.325){$t= \SI{14}{\second}$};
    \node[font = {\small}] at (0.62,0.26){$t= \SI{15}{\second}$};
     % \node[font = {\small}] at (0.86,0.19){$t= \SI{16}{\second}$};
     % \node[] at (0.495,0.6){$t= \SI{7}{\second}$};

     % \node[] at (0.41,0.2){$t= \SI{7}{\second}$};
     \node[font = {\small}] at (0.34,0.18){$t= \SI{5}{\second}$};
     % \node[font = {\small}] at (0.16,0.19){$t= \SI{4}{\second}$};
     \node[font = {\small}] at (0.09,0.19){$t= \SI{3}{\second}$};
\end{scope}

% \node[inner sep=0pt] at (0,0)
%     {\includegraphics[width=\textwidth]{figures/1agents.png}};
% \node[] at (-1,-0.5){$t= \SI{5}{\second}$};
\end{tikzpicture}

% \end{document}

%% file: figures/tjunc3.tex
% \documentclass{article}
% %
% %% Einige Packages, die nützlich sind für die Erstellung von Grafiken:
% \usepackage[utf8]{inputenc}
% \usepackage{soulutf8}
% \usepackage{pgfplots}
% \usepackage{siunitx}
% \usetikzlibrary{plotmarks}
% \usepgfplotslibrary{groupplots}
% \usepgfplotslibrary{statistics}
% \usepgfplotslibrary{external}
% % \tikzexternalize
% % Für das Einbinden von Tabellen in Bilder können diese Packages hilfreich sein:
% %\usepackage{harveyBalls}
% %\usepackage{fontawesome}
% %\usepackage{multirow}
% % \usepackage[per-mode=symbol]{siunitx-v2}
% \begin{document}
% 	\definecolor{TUMGreen}{RGB}{162, 173,0}
% 	\definecolor{TUMBlue}{RGB}{0,101,189}
% 	\definecolor{TUMOrange}{RGB}{227,114,34}

\begin{tikzpicture}%[trim axis left, ,trim axis right]

\node[anchor=south west,inner sep=0] (Bild) at (0,0) {\includegraphics[width=0.29\linewidth]{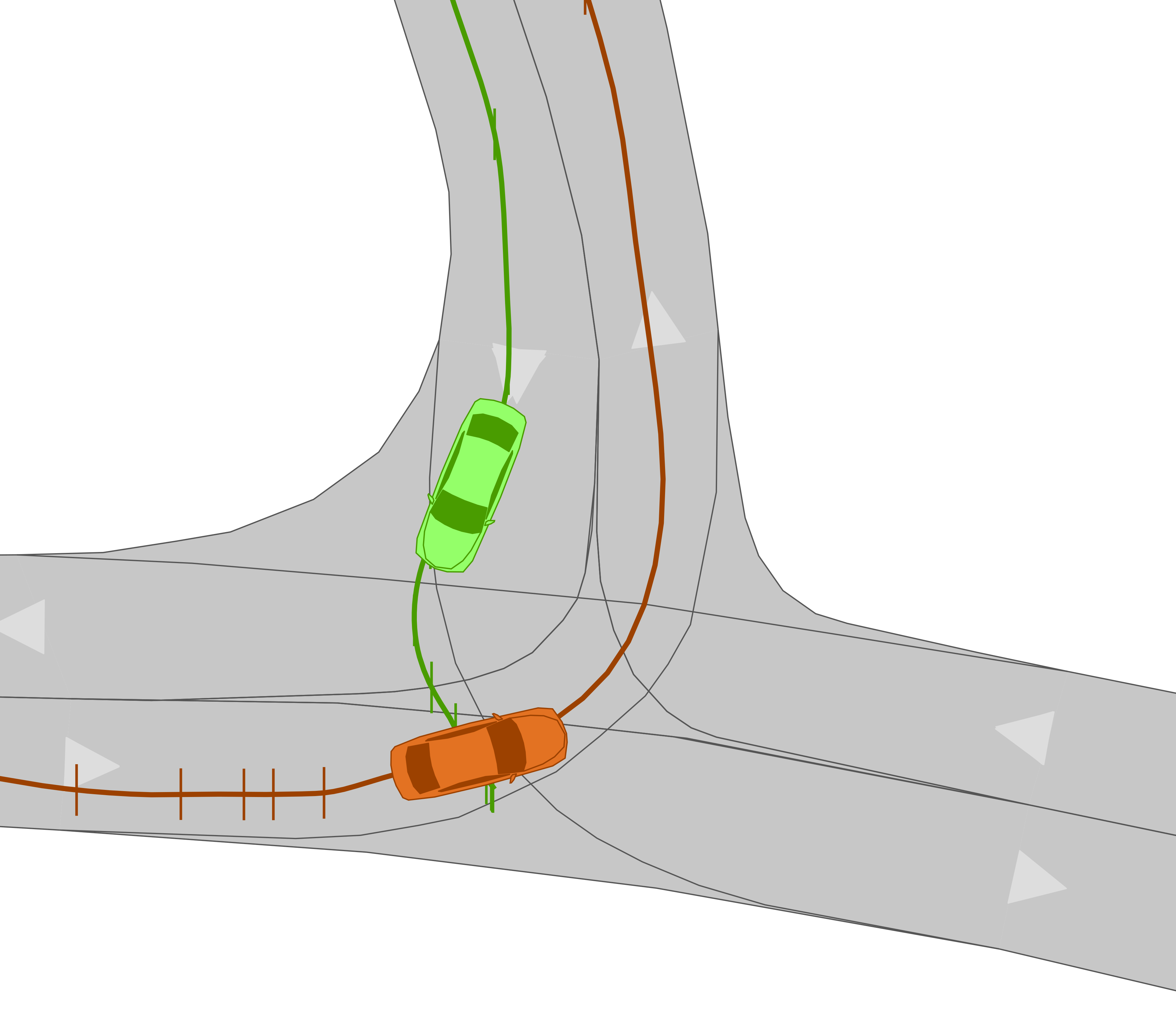}};
\begin{scope}[x=(Bild.south east),y=(Bild.north west)]
     \node[font = {\small}] at (0.53,0.87){$t= \SI{4}{\second}$};
     % \node[font = {\small}] at (0.485,0.72){$t= \SI{5}{\second}$};
     % \node[] at (0.49,0.64){$t= \SI{6}{\second}$};
     % \node[font = {\small}] at (0.55,0.58){$t= \SI{7}{\second}$};
      \node[font = {\small}] at (0.54,0.51){$t= \SI{8}{\second}$};

    % \node[font = {\small}] at (0.445,0.45){$t= \SI{9}{\second}$};
    % \node[font = {\small}] at (0.5,0.39){$t= \SI{10}{\second}$};
    % \node[font = {\small}] at (0.425,0.33){$t= \SI{11}{\second}$};
    
    % \node[font = {\small}] at (0.09,0.185){$t= \SI{3}{\second}$};
     \node[font = {\small}] at (0.18,0.18){$t= \SI{4}{\second}$};
    % \node[font = {\small}] at (0.215,0.185){$t= \SI{5}{\second}$};
    % \node[font = {\small}] at (0.3,0.185){$t= \SI{7}{\second}$};
     \node[font = {\small}] at (0.59,0.26){$t= \SI{8}{\second}$};
     % \node[font = {\small}] at (0.61,0.445){$t= \SI{9}{\second}$};

     \node[font = {\small}] at (0.63,0.99){$t= \SI{10}{\second}$};
\end{scope}

% \node[inner sep=0pt] at (0,0)
%     {\includegraphics[width=\textwidth]{figures/1agents.png}};
% \node[] at (-1,-0.5){$t= \SI{5}{\second}$};
\end{tikzpicture}

% \end{document}

%% file: data/diff_tjunc2.tex
%%%% over t
\begin{tikzpicture}[trim axis left, ,trim axis right]

	\begin{axis}[
		boxplot/draw direction=y,
		width=.95\linewidth,
		height=0.5\linewidth,
        xmin = 2, xmax = 15.0,
        ymax = 16,        
        legend style={cells={align=left}},
%		symbolic x coords={Acceleration},
		xtick={3,6,9,12},
		ytick={4, 8,12},
		minor x tick num=1,
        % minor y tick num=1,
		axis x line = bottom,
		axis y line = left,
        ylabel style={align=center},
		ylabel={v (\si[per-mode=reciprocal]{\metre\per\second})},
		xlabel={t (\si[per-mode=reciprocal]{\second})},
 legend style={draw=none},
 legend columns=2, 
 % legend cell align = left
	legend style= {at={(0.5,1)},anchor=south, nodes={scale=0.85, transform shape},
		legend cell align = left, /tikz/column 2/.style={
                column sep=5pt}},
	]

    \addplot [thick, draw=TUMOrange,dashed,] 
            table[x = t, y = 60000_0_v, col sep=comma] {data/diff_tjunc.csv};
    \addlegendentry{Veh. 1, agent}
	\addplot [thick, draw=TUMGreen,dashed ] 
            table[x = t, y = 5_0_v, col sep=comma] {data/diff_tjunc.csv};
            \addlegendentry{Veh. 2, deterministic}
            
    \addplot [thick, draw=TUMOrange, dotted ] 
            table[x = t, y = 60000_idm_v, col sep=comma] {data/diff_tjunc.csv};
            \addlegendentry{Veh. 1, agent}
	\addplot [thick, draw=TUMGreen, dotted] 
            table[x = t, y = 5_idm_v, col sep=comma] {data/diff_tjunc.csv};
            \addlegendentry{Veh. 2, semi-deterministic}

    \addplot [thick, draw=TUMOrange, ] 
            table[x = t, y = 60000_2_v, col sep=comma] {data/diff_tjunc.csv};
            \addlegendentry{Veh. 1, agent}
	\addplot [thick, draw=TUMGreen ] 
            table[x = t, y = 5_2_v, col sep=comma] {data/diff_tjunc.csv};
            \addlegendentry{Veh. 2, agent}

\end{axis}

\end{tikzpicture}